\documentclass{article}
\usepackage{spconf,amsmath,graphicx}

\usepackage{ragged2e}
\usepackage{amsmath}
\usepackage{graphicx}
\usepackage{algorithm}
\usepackage{hyperref}
\usepackage{color}
\usepackage[noend]{algpseudocode}
\usepackage{bm}
\usepackage{lipsum}
\usepackage{amsfonts}
\usepackage{verbatim}
\usepackage[export]{adjustbox}
\usepackage{caption}

\title{Sequence Training of DNN Acoustic Models With Natural Gradient}
\name{ Adnan Haider  \&  Philip C. Woodland   \thanks{Adnan Haider has been funded by the IDB Cambridge International Scholarship and partly supported by the EPSRC Programme Grant EP/I031022/1 (Natural Speech Technology).}  }
\address{Cambridge University Engineering Dept., Trumpington St., Cambridge, CB2 1PZ U.K.\\ Email:
{\{mah90, pcw\}@eng.cam.ac.uk}}

\begin{document}
\ninept
\maketitle
 \begin{abstract}
Deep Neural Network (DNN) acoustic models often use discriminative sequence training  that optimises an objective function that better approximates the word error rate (WER) than frame-based training. Sequence training is normally implemented using Stochastic Gradient Descent (SGD) or  Hessian Free (HF) training. This paper proposes an alternative batch style optimisation framework that employs a Natural Gradient (NG) approach to traverse through the parameter space. By correcting the gradient according to the local curvature of the KL-divergence, the NG optimisation process converges more quickly than HF. Furthermore, the proposed NG approach can be applied to any sequence discriminative training criterion. The efficacy of the NG method is shown using experiments on a Multi-Genre Broadcast (MGB) transcription task that demonstrates both the computational efficiency and the accuracy of the resulting DNN models.
\end{abstract}


\section{Introduction}
\label{sec:intro}
With the advent of GPU computing and careful network initialisation \cite{Bengio2007a}, a hybrid of Deep Neural Networks (DNNs) combined with Hidden Markov models (HMMs) have been shown to outperform traditional Gaussian Mixture Model (GMM) HMMs  on a wide range of speech recognition tasks \cite{Hinton2012}. DNNs, due to their deep and complex structure,  are better equipped to model the underlying nonlinear data manifold in contrast to GMMs. However, the existence of such structures also creates complex dependencies between the model parameters which can make such models  difficult to train with standard Stochastic Gradient Descent (SGD). 

The task of finding the best procedure to train DNNs is an  active area of research that has been rendered more challenging by the availability of ever larger  datasets.  It is therefore necessary to develop techniques that lead to rapid convergence and are also stable in training.
For example, second order optimisation methods can help to alleviate the problems of vanishing and exploding gradients by rescaling the gradient direction to adjust for the high non-linearity and ill-conditioning of the objective function. For convex optimisation problems \cite{Bottou2016}, such methods have been shown  to improve  the  convergence of both batch and stochastic methods.  Martens \cite{Martens2010} was one of the first researchers to  successfully apply second order  methods to train DNNs. Instead of rescaling the gradients with a `local' curvature matrix directly, the author advocated using a  `Hessian-Free' (HF) approach which generates an approximate update by the iterative Conjugate Gradient (CG) algorithm.  Such a method has been shown to be effective for lattice-based discriminative sequence training \cite{Kingsbury2012} when applied with large batch sizes. When training on  speech data, it is advantageous to randomly select the frames in each mini-batch when using SGD training with the frame-based Cross Entropy (CE) objective function.
However, for lattice-based discriminative sequence training, information from a forward-backward pass through the complete lattice of individual utterances is required to be calculated.  Hence, data randomisation can only be effectively performed at the utterance level.
 In this context, \cite{Sainath2013c} found that second order methods like HF that accumulate gradients over large batches, perform better than utterance-based mini-batch SGD.

However, as an optimiser, the HF method is slow and suffers from drawbacks. In a stochastic HF approach, where we employ only a single batch to compute the gradient estimate, the optimisation process has been seen to often converge to a sub-optimal solution. To address these drawbacks, in  \cite{Dognin2013}, the authors  proposed combining the HF optimiser with the Stochastic Average Gradient (SAG), where at each iteration,  the  CG  algorithm is initialised with a weighted sum of the current and previous gradient estimates. On a large amount of data, the proposed modification was shown to improve both the speed and convergence of  HF.  However, when faced with moderate sized training sets, we have found  that such an approach can suffer from over-fitting due to the  mismatch of training criteria between the DNN sequence discriminative  objective function and  the true objective which is to reduce the Word Error Rate (WER).

In this work, we propose an alternative optimisation framework for sequence training that is more effective with large batch sizes than HF. Instead of minimising a local second order model, it is shown that substantial improvements in both training speed and convergence 
can result from using  CG to derive an approximation to the Natural Gradient (NG)  direction. In NG, the update direction chosen at each iteration is the product of the inverse of the empirical Fisher Information (FI) matrix with the gradient estimate. Essentially this amounts to performing gradient descent on the space of densities $p_{\bm{\theta}}(\mathcal{H},\mathcal{O})$, captured by different parameterisations of $\bm{\theta}$.   In Automatic Speech Recognition (ASR), this approach has previously been applied for frame-based classification  under a different optimisation framework. In \cite{Povey2014},  the authors assume a block diagonal structure for the FI matrix and computes a stochastic estimate of the NG direction directly. In contrast, this paper makes no such assumptions about the structure of the empirical FI matrix and computes an approximation to the NG direction by running  a few iterations of CG. Unlike the framework described in \cite{Povey2014}, our optimisation framework has the flexibility that it can be used to train DNNs with respect to any sequence classification criterion.

This paper is organised as follows. Sec.~\ref{sec:background} provides a brief review of HMM-DNN systems in ASR. Sec.~\ref{sec:ng} introduces NG and provides a derivation of the empirical FI matrix for use in sequence training. In Sec.~\ref{sec:2ndorder},  a slight detour is taken to discuss 2nd order methods and discuss the  model that is minimised at each iteration. In Sec. 5,  NG is formulated for sequence training within the same framework as HF. Sec.~\ref{sec:cg} discusses CG while in Sec.~\ref{sec:ngmbr},  it is shown how NG improves MBR training. The experimental setup for ASR experiments on Multi Genre Broadcast (MGB) data is presented in Sec.~\ref{sec:setup}, results in Sec.~\ref{sec:results}, followed by conclusions.

\section{Background}
\label{sec:background}
 
Large Vocabulary Continuous Speech Recognition (LVCSR) is an application of sequence to sequence modelling, where given a continuous acoustic waveform $\mathcal{O}^r = \{ \bm{o}^r_1,\bm{o}^r_2 \cdots \bm{o}^r_T\}$, the objective is to generate its most likely word sequence, $\mathcal{H}= \{ \bm{w}^r_1,\bm{w}^r_2 \cdots \bm{w}^r_L\}$. The problem can be essentially cast as a general problem of supervised learning where given some seen examples  $\{\mathcal{O}^r, \mathcal{H}^i\}_{r=1}^N$, the goal is to learn the relationship between the input space $\mathcal{X}$ and the output space $\mathcal{Y}$ from the training data. In parametric inference,  we model this relationship by  a statistical model $P_{\bm{\theta}}(\mathcal{H}|\mathcal{O})$.  For speech processing and a range of other sequential modelling tasks, this currently normally corresponds to using HMMs \cite{Gales2008} with output scaled likelihoods computed using DNNs. HMMs  belong to the  class of generative discrete latent variable graphical models that aim to model  $P(\mathcal{H}|\mathcal{O})$ by modelling the joint distribution  $P(\mathcal{O},\mathcal{H})$ instead.  Using bayes theorem, the posterior probabilities can then computed as $ P(\mathcal{H} |\mathcal{O}) = p(\mathcal{O},\mathcal{H} )/ {p(\mathcal{O})}$. Note that  any of the quantities in this expression   can be obtained by either marginalising or conditioning with respect to the appropriate variables in the joint distribution.

In ASR, the training of DNNs is normally conducted in  two phases: frame-based training followed by sequence training. The first phase trains the  DNN model parameters w.r.t the CE criterion, an objective function that is particularly suited for frame-level classification. 
In contrast, the second phase, employs a  sequence discriminative criterion to further train the  model.  Since ASR is a sequence to sequence modelling task, having this pipeline has been found to achieve  reductions in the WER \cite{Kingsbury2009}.  At present, for  sequence discriminative training, two popular classes of  objective functions are used:
\begin{enumerate}
\item Maximum Mutual Information (MMI) \cite{Woodland2002} which maximises the mutual information between the word sequence and the  information extracted by the recogniser from the given observation sequence.
\begin{align}
F_{\rm{MMI}} (\bm{\theta}) =  \frac{1}{R} \sum_{r=1}^R  \mbox{ log } P_{\bm{\theta}}(\mathcal{H}^r |\mathcal{O}^r)   \label{equ1}
\end{align}
 This also includes the popular `boosted' variant \cite{Povey2008} which is inspired by margin-based objective functions.

\item Minimum Bayes' Risk (MBR) \cite{Gibson2006}  corresponds to a family of methods which  minimises the  average expected loss computed over the hypothesis space  for a given observation sequence when the posterior distribution $P(\mathcal{H} |\mathcal{O})$ is modelled by  $P_{\bm{\theta}}(\mathcal{H}|\mathcal{O})$. This corresponds to the following objective function:
\begin{align}
F_{\rm{MBR}} (\bm{\theta}) =  \frac{1}{R} \sum_{r=1}^R \sum_{\mathcal{H}'} ,P_{\bm{\theta}} (\mathcal{H}' |\mathcal{O}^r)   L(\mathcal{H}',\mathcal{H}^r) \label{equ2}
\end{align}
\end{enumerate}
Here   $ L(\mathcal{H}',\mathcal{H}^r) $ denotes the mismatch between the proposed hypothesis and the correct hypothesis. To assist generalisation, this  is often computed at a finer level of granularity such as at  the phone or `physical' context-dependent state level. For string sequences of arbitrary length,   the Levenshtein distance \cite{Levenshtein1966} is the usual metric of measuring the mismatch between them. However, the cost of employing such a metric increases with the length of the sequences. Due to this computational overhead, in practice, we annotate arcs that we follow in the recognition trellis with a local loss. When this local loss assignment is done at the phone level, this corresponds to the Minimum Phone Error (MPE) objective function \cite{Povey2002}\footnote{In practice, in MPE training,  the objective function is actually maximised. This is because instead of assigning a local loss to each phone  arc $q$, the local $phoneAcc(q)$ is used. However, by negating this accuracy we can cast  this method as an MBR risk function.}. While, when the loss assignment is done at the physical state level, this is the state-level MBR (sMBR) objective function.


\section{Natural Gradient}
\label{sec:ng}
One particular approach for training DNNs that has recently experienced renewed popularity is the method of NG\cite{Pascanu2013a,Desjardins2015,Martens2014}. This method was first proposed by Amari \cite{Amari1998a}, as an effective optimisation method for training parametric density models.   Instead of formulating gradient descent in the  Euclidean parameter space, we establish a geometry  on the space of density functions and perform steepest descent on that space.  Amari showed \cite{Amari1997}, that by inherently doing optimisation on the space of density models,  these models can be trained in a more effective way than with standard gradient descent.  This work extends its application to the domain of sequence training.  In this section,  the geometry is developed that is needed to traverse through the functional manifold of  distributions $P_{\bm{\theta}}(\mathcal{H}|\mathcal{O}) $.

\subsection{Fisher Information Matrix for sequence to sequence modelling}

Let $\mathcal{F}$ be the family of densities $p_{\bm{\theta}}(\mathcal{H}|\mathcal{O})$ that can be captured by different parameterisations of $\bm{\theta}$. To formulate gradient descent on the functional manifold $\mathcal{F}$, it is necessary to quantify how the density  $p_{\bm{\theta}}(\mathcal{H}|\mathcal{O})$ changes when one adds a small quantity $\Delta \bm{\theta}$ to $ \bm{\theta}$. For probability distributions, such changes in behaviour can be captured by the KL-divergence, ${\rm KL}\left (p_{\bm{\theta}}(\mathcal{H}|\mathcal{O}) \|p_{\bm{\theta}+\Delta \bm{\theta}}(\mathcal{H}|\mathcal{O})\right)$.  The KL-divergence is  a functional that maps the space of  distributions  $\mathcal{F}$ to $\mathbb{R}$.  Since each distribution itself is a function of $\bm{\theta}$, the KL-divergence can also be expressed as a smooth function of $\bm{\theta}$  through compositionality. Using a Taylor approximation, within a  convex neighbourhood of any  given point $\bm{\theta}_k$, its behaviour can be bounded  as follows:
\begin{align}
&  {\rm KL} \left (p_{\bm{\theta}}(\mathcal{H}|\mathcal{O}) \|p_{\bm{\theta}+\Delta \bm{\theta}}(\mathcal{H}|\mathcal{O})\right)
  \notag \\
\approx & - \Delta \bm{\theta}^T E_{p_{\bm{\theta}}(\mathcal{H}|\mathcal{O})} \left [ \nabla \mbox{ log }p_{\bm{\theta}}(\mathcal{H}|\mathcal{O}) \right]  \notag\\
& - \frac{1}{2} \Delta \bm{\theta}^T  E_{p_{\bm{\theta}}(\mathcal{H}|\mathcal{O})} \left [  \nabla^2  \mbox { log }  p_{\bm{\theta}}(\mathcal{H}|\mathcal{O})\right ]\Delta \bm{\theta}  \label{equ3}
\end{align}

Using the fact that each candidate  $p_{\bm{\theta}}(\mathcal{H}|\mathcal{O})$  is a valid probability distribution, it can be shown that the first term  in the quadratic approximation $E_{p_{\bm{\theta}}(\mathcal{H}|\mathcal{O})} \left [ \nabla \mbox{ log }p_{\bm{\theta}}(\mathcal{H}|\mathcal{O}) \right] $ equates to zero. This allows the KL divergence  to be locally approximated by the bilinear form:

\begin{align}
& {\rm KL}\left (p_{\bm{\theta}}(\mathcal{H}|\mathcal{O}) \|p_{\bm{\theta}+\Delta \bm{\theta}}(\mathcal{H}|\mathcal{O})\right)   \notag \\
 &\approx - \frac{1}{2} \Delta \bm{\theta}^T  E_{p_{\bm{\theta}}(\mathcal{H}|\mathcal{O})} \left [  \nabla^2  \mbox { log }  p_{\bm{\theta}}(\mathcal{H}|\mathcal{O})\right ]\Delta \bm{\theta} \label{equ4}
\end{align}

For inference models $P_{\bm{\theta}}(\mathcal{H}|\mathcal{O})$, the Fisher Information  $I_{\bm{\theta}} $,  for the random variable $\bm{\theta}$ is the expected outer product of the likelihood  \textbf{score}:
\begin{align}
I_{\bm{\theta}}  &=  E_{p_{\bm{\theta}}(\mathcal{H}|\mathcal{O})}  \left [ \left( \nabla \mbox { log }  P_{\bm{\theta}}(\mathcal{H}|\mathcal{O})\right ) \left( \nabla \mbox { log }  P_{\bm{\theta}}(\mathcal{H}|\mathcal{O})\right )^T \right ] 
 \label{equf}    
\end{align}

 In scenarios where  ({\ref{equ4}}) holds, $I_{\bm{\theta}}$  can be shown to be equal to the negative of the expectation of the Hessian w.r.t the distribution $p_{\bm{\theta}}(\mathcal{H}|\mathcal{O})$. Thus by substituting (\ref{equf})  into (\ref{equ4}), we can now locally approximate  the KL divergence by the inner product:
\begin{align}
{\rm KL }\left (p_{\bm{\theta}}(\mathcal{H}|\mathcal{O}) \|p_{\bm{\theta}+\Delta \bm{\theta}}(\mathcal{H}|\mathcal{O})\right) \approx \frac{1}{2} \Delta \bm{\theta}^T I_{\bm{\theta}}  \Delta \bm{\theta} \label{equ7}
\end{align}

Instead of computing the expectation of (\ref{equf}) directly, in this work we approximate  $I_{\bm{\theta}}$ with the following Monte-Carlo estimate:
\begin{align}
 \hat{I}_{\bm{\theta}}  =  \frac{1}{R} \sum_{r=1}^R \left [ \left( \nabla \mbox { log }  P_{\bm{\theta}}(\mathcal{H}^r|\mathcal{O}^r)\right ) \left( \nabla \mbox { log }  P_{\bm{\theta}}(\mathcal{H}^r|\mathcal{O}^r)\right )^T \right ]  \label{equ8}
\end{align}
The matrix  $ \hat{I}_{\bm{\theta}}$ is the empirical FI matrix and corresponds to  the outer product of the Jacobian of the MMI criteria.  By assigning an inner product of the form $ \hat{I}_{\bm{\theta}}$ which is dependent $\bm{\theta}$, we have assigned a Riemann metric on the functional manifold $\mathcal{F}$. Such a formulation now allows us to  define a  notion of a local  `distance'  measure in the space of joint distributions $p_{\bm{\theta}}(\mathcal{H}|\mathcal{O}) $.


\section{Second order optimisation} 
\label{sec:2ndorder}
Before proceeding to formulate NG for sequence training,  a slight detour is taken to discuss how  methods like HF minimise a  given objective criterion. Assuming that the objective criterion $F(\bm{\theta})$ is sufficiently smooth, for a given point $\bm{\theta}_k$ in the parameter space, there exists an open convex neighbourhood where its behaviour can be locally approximated as:
\begin{align}
F(\bm{\theta}_k + \Delta \bm{\theta}) \approx  F(\bm{\theta}_k ) + \nabla F(\bm{\theta}_k ) \Delta \bm{\theta} + \frac{1}{2} \Delta \bm{\theta}^T H  \Delta \bm{\theta} \label{equ9}
\end{align}
Instead of minimising $F(\bm{\theta})$,  second order methods aim to minimise this approximate quadratic at each iteration.
 
\subsection{Approximating the Hessian with the Gauss-Newton matrix}
 Within the convex neighbourhood of  $\, the cbm{\theta}_k$ritical point $\Delta \bm{\theta} = H^{-1}\nabla F(\bm{\theta}_k )$  corresponds to a unique minimiser only when the Hessian  $H$ is positive definite.  However, when we restrict our choice of predictive functions $\mathcal{F}$ to DNN models, the optimisation problem is no longer guaranteed to be convex. This means that within a convex neighbourhood of  $\bm{\theta}_k$, the Hessian associated with the 2nd order Taylor model is no longer guaranteed to be positive definite. In such a scenario, minimisation of ({\ref{equ9})  no longer guarantees a decrement in the true objective function. To overcome this issue, the general practice is to replace the 2nd order model with an approximate model in which the Hessian is replaced by a positive definite matrix $B$.  

\begin{align}
\hat{F}(\bm{\theta}_k + \Delta \bm{\theta}) \approx  F(\bm{\theta}_k ) + \nabla F(\bm{\theta}_k ) \Delta \bm{\theta} + \frac{1}{2} \Delta \bm{\theta}^T B  \Delta \bm{\theta} \label{equ10}
\end{align}
In the work by  Sainath and Kingsbury \cite{Sainath2013c}, the authors have claimed to achieve  reductions in WER on large datasets  when they employ a Gauss Newton (GN) approximation of the Hessian matrix.  For convex loss functions defined on DNN output activations, this approximation always yields a positive semi definite matrix. In neighbourhoods where the neural network objective   $F(\bm{\theta})$ can be locally modelled by its first order approximation, it can be shown that the GN  matrix represents the Hessian exactly. 

The GN  matrix takes the form   $ J_t ^{T}\nabla^2 L_tJ_t$ where $J_t$ is the Jacobian of the linear output activations w.r.t $\bm{\theta}$ and   $\nabla^2 L_t$ is the  Hessian of the  loss with respect to the DNN linear output activations at time $t$. For MBR loss functions,  $\nabla^2 L_t$  takes the  following form:
\begin{align}
\nabla^2 L_t &= \frac{\kappa^2}{R} \left [\mbox{diag}( \hat{\bm {\gamma}}^r_t) -  \hat{\bm{\gamma}}^r_t (\bm{\gamma}^r_t)^T \right] \label{equ11}
\end{align}
with $ \hat{\bm{\gamma}}^r_t=  \bm{\gamma}^r_t \odot \bm{L}(\bm{s}) $. Here, 
 
\begin{itemize}
\item $\bm{L}(\bm{s})$ is a vector whose $i$th entry  corresponds to the difference between $\check{L}(i)$, the  posterior weighted sum of the local losses computed over all the lattice paths that pass through arcs containing the state $i$, and  
$c_{\rm{avg}}^L$, the posterior weighted sum of the loss computed over all the lattice paths.
\item  $\bm{\gamma}^r_t$  is a vector whose entries correspond to   the posterior  probability associated with the states (DNN output nodes) at time $t$ within the consolidated lattice.
\item $\kappa$  is the acoustic scaling factor.  To assist generalisation, when propagating through the trellis, acoustic probabilities are often raised to a power less than unity to allow  less likely sentences to  become more important during discriminative sequence training.

 \end{itemize}
In the next section, we show that we can formulate NG in a similar way where  at each iteration we minimise a function of the  form  of (\ref{equ10}).


\section{Formulating Natural Gradient for Sequence Training}
Having established a notion of a local distance on the space of discriminative distributions $\mathcal{F}$ yielded by a hybrid HMM-DNN models,  an algorithm is required that minimises a functional  $L : p_{\bm{\theta}}(\mathcal{H}| \mathcal{O}) \in \mathcal{F}  \rightarrow   \mathbb{R}$. 
In Sec.~\ref{sec:background}, it has been suggested that this can be achieved by  either using  one of  the  MBR  loss functions or the  negation of  the MMI criterion.
Each iteration of a typical iterative optimisation algorithm computes an  iterate   $p_{\bm{\theta}_{k+1}}(\mathcal{H}|\mathcal{O}) $ on the basis of information pertaining to the current iterate $p_{\bm{\theta}_{k}}(\mathcal{H}|\mathcal{O}) $. Since  the Riemannian geometry  describes only the local behaviour around  $p_{\bm{\theta}_{k}}(\mathcal{H}|\mathcal{O}) $, to generate a candidate function,  the following greedy strategy is formulated:
\begin{align}
\hat{p}_{\bm{\theta}_{k+1}}(\mathcal{H},\mathcal{O}) = \arg\min_{ p_{\bm{\theta}_{k+1}} \in \mathcal{F}}  F (p_{\bm{\theta}_{k+1}})   & \mbox{  s.t  }  {\rm KL} \left (p_{\bm{\theta}} \|p_{\bm{\theta}+\Delta \bm{\theta}}\right)\leq \epsilon_k 
\end{align}
where  we  use $F$ to represent  the chosen sequence classification criterion. 

Using (\ref{equ7}) and (\ref{equ8}), this can  be reformulated as an optimisation problem in the parameter space:

\begin{align}
\bm{\theta}_{k+1} =  \arg\min_{\bm{\theta}} F(\bm{\theta})  \mbox{  s.t  }  \frac{1}{2}( \bm{\theta} - \bm{\theta}_k )^T \hat{I}_{\bm{\theta}_k} ( \bm{\theta} - \bm{\theta}_k ) \leq \epsilon_k 
\end{align}
In this work, the probing of the parameter space is restricted to within a convex neighbourhood of $\bm{\theta}_k$ where  a first order approximation to $ F(\bm{\theta})$ can be used: $ F(\bm{\theta}) \approx  F(\bm{\theta}_k)  + \nabla  F(\bm{\theta}_k) ^T\Delta \bm{\theta} $. In doing so, the optimisation problem  can now be cast  as a first order minimisation problem with in a trust region. Since the minimum of the first order approximation  is at infinity, the optimisation dynamics will always jump to the border of the trust region. Thus, at each iteration  the following constrained optimisation problem is solved:

\begin{align}
F(\bm{\theta}_k + \Delta \bm{\theta}) \approx  F(\bm{\theta}_k ) + \nabla F(\bm{\theta}_k ) \Delta \bm{\theta} + \frac{\lambda}{2} \Delta \bm{\theta}^T \hat{I}_{\bm{\theta}_k}   \Delta \bm{\theta}
\end{align}
Hence it can be seen that under such a formulation,  at each iteration of NG, a function of the form of (\ref{equ10}) is minimised.


\section{Conjugate Gradient}
\label{sec:cg}
Differentiating (\ref{equ10}) and setting it to zero yields the Newton direction $ \Delta \bm{\theta} = B^{-1} \nabla F(\bm{\theta}) $, where $B$  now corresponds to either GN or $\hat{I}_{\bm{\theta}_k}$. However, the Newton direction does not scale well with the dimension $D$ of the optimisation problem. Computing this direction directly is  expensive  in terms of both computation and storage as  storing  $B$ requires $\mathcal{O}(D^2)$ storage and inverting it incurs a cost of $\mathcal{O}(D^3)$. These obstacles however, can be overcome if we employ \textbf{ inexact} Newton methods. In particular, rather than computing the Newton direction  exactly through matrix factorisation techniques,  we solve the system $B \Delta \bm{\theta} =  \nabla F(\bm{\theta}) $ using the iterative linear Conjugate Gradient(CG) algorithm \cite{Shewchuk1994b}. When B represents the  GN matrix, such a method is called Hessian-Free.

 
 Like many iterative linear system techniques, CG applied to (11) does not require access to $B$ itself but only its vector products. Pearlmutter \cite{Pearlmutter1994}  has shown  that such  matrix-vector products can be computed efficiently  through  appropriate modifications of the forward and backward pass. By iteratively solving the linear system,  the CG algorithm minimises the local quadratic  of (\ref{equ10}) by continuously improving  the search direction it initially starts with.  In our implementation,  CG is initialised with an estimate of the gradient. It can be shown that if this estimate is accurate then  CG in its initial iterations  will be able to find updates that immediately  improves upon this direction.  This proves to be quite an advantage, especially when there is hard limit of the maximum number of allowed  iterations.
 
     Within this  framework, using only a  small number of CG iterations is also desirable for computational reasons.  Within each iteration of CG, the computation of each matrix-vector product is as expensive as a gradient evaluation.  This means that  as the size of the dataset increases, the cost of running a single CG iteration also increases linearly. To address this issue, in practice, a subsampled approximation to the candidate $B$ is used when  (\ref{equ10}) is minimised. It was found that this approach is actually quite reasonable,  as the training  iterations were found to be more tolerant to noise in the $B$'s estimate than it  to the noise in the gradient estimate.  Following Kingsbury's approach \cite{Kingsbury2012}, our default recipe consists of sampling 1\% of the  training  set to do matrix-vector products for each run of CG.   Effectively, by using such a  setup, an optimisation algorithm has been framed where the CG iterations contribute only a small percentage of the total cost and the gradient computations are  inherently data parallel.


\section{How Natural Gradient improves MBR training}
\label{sec:ngmbr}
 In ASR, the true evaluation criteria is WER, a function which is  not differentiable.  In MBR training,  a differentiable loss function that is a smoothed approximation of the WER is minimised (eg. MPE \cite{Povey2002}  or sMBR \cite{ Gibson2006} ).  Since such loss functions are only approximations of the WER, during training, over-fitting can also occur due to the mismatch in training criterion. In this section, it is shown how rescaling the MBR gradients with the inverse of the empirical FI matrix can ensure that minimising an MBR criterion correlates well with achieving reductions in the WER.

In Sec.~\ref{sec:background}, it was highlighted that to achieve better generalisation, the mismatch between $(\mathcal{H}',\mathcal{H}^r)$ is often computed at a finer level of granularity, such as at phone or `physical' state level.  The mismatch between string sequences of arbitrary length is usually computed by the  Levenshtein distance \cite{Levenshtein1966}. However, the cost of employing such a metric increases with the length of the sequences. Due to this computational overhead, in practice, arcs  in the recognition trellis are annotated  with a local loss\footnote{In MPE training, this is done at the level of a phone  while in sMBR training, this is done at the level of `physical' states.}.  Then during training,  a weighted sum of these individual local losses is minimised, where the weights correspond to the posterior probability of being on an individual arc.   
 
  \begin{align}
F_{\rm{MBR}} (\bm{\theta}) &=  \sum_{r=1}^R  \frac{1}{Z} \sum_{q} \alpha_q \beta_q L(q,q^r)
\end{align}
 where
\begin{itemize}
\item $ \alpha_q $ represents the  forward probability traversing through arc $q$ in the recognition lattice. This  in turn can be un-rolled in a recursive fashion:

$\alpha_q  = \sum_{r \mbox{ preceding } q} \alpha_r t_{rq}^\kappa p({O}^{r,q}| M_{q} ,\bm{\theta})^\kappa$ 

 $p(O^{r,q}| M_{q})$  here represents the arc likelihood and  $O^{r,q}$ corresponds to the frames observed  between the start and end times of the arc $q$.
 
\item $\beta_q $ represents the  backward probability  of traversing from the  arc $q$ to the end of the recognition trellis. Like the forward probability, this too can be recursively defined as:

$\beta_q   =  \sum_{s \mbox{ following } q}  t_{qs}^\kappa p({O}^{r,s}| M_{s},\bm{\theta})^\kappa \beta_s$
\item  the partition function $Z$ represents  the sum of joint probabilities of transversing through all paths through the lattice.
\end{itemize}
In a lattice-based framework,  the individual local losses are initially computed using a CE trained model  and do not change during the course of training. What does change are the weights we assign to them. For arcs where $ L(q,q^r)$ is high, at each iteration, we adapt the model parameters to decrease $p({O}^{r,q}| M_{q} ,\bm{\theta})$ which in (16) can be seen  to reduce the weights associated with these arcs.  Under both HF and  NG, at each iteration, an update is generated  by re-scaling the gradient direction:
\[\Delta \bm{\theta} = B^{-1} \nabla F_{\rm{MBR}} (\bm{\theta})\]
 Since by definition $B$ is real and symmetric, under a change of basis, the above expression reduces to:
 \begin{align}
\Delta{\bm{\theta}} = \sum_{i} \frac{1}{\hat{\mu}_i} \left( \bm{v}_i^T  \nabla F_{\rm{MBR}} \right) \bm{v}_i
\end{align}
Here the basis vectors $\{ \bm{v}_i \}_i$ correspond to the eigen vectors of the matrix. Under this formulation, it can be seen how the choice of the matrix B impacts the steps taken by the optimiser in the parameter space. In HF,  the eigen values $\hat{\mu}_i$ encode the curvature of the loss function w.r.t  the DNN output activations.  The HF optimizer uses this information to rescale the steps it takes in each eigen direction. However, since GN does not encode the curvature  w.r.t the model parameters (whose optimisation is the main objective), it  is not always guaranteed for such scalings to be always optimal.

In contrast,  when  $B$ corresponds to $\lambda \hat{I}_{\bm{\theta}}$, the scaling factor is  $\hat{\mu_i} = \lambda \mu_i$, where $\mu_i$ is the respective eigenvalue associated with direction $\bm{v}_i$.  Under such rescaling, the optimisation process will now move more cautiously along directions  which have a significant impact on  $P_{\bm{\theta}}(\mathcal{H}^r |\mathcal{O}^r)$ of each utterance $r$. With respect to MBR training, this tries to ensure that  the probability $p({O}^{r,q^r}| M_{q^r} ,\bm{\theta})$ associated with the reference arcs does not change while the model parameters are  adapted to reduce  $p({O}^{r,q}| M_{q},\bm{\theta})$ of the hypotheses arcs.   The Lagrange multiplier $\lambda$ controls the leverage  between minimising the objective and enforcing this constraint. Through suitably adjusting $\lambda$, the optimisation process will follow a path where minimising the MBR criterion correlates well with achieving reductions in WER.

\section{Experimental Setup}
\label{sec:setup}

The  DNN training techniques were evaluated  on data from the  2015 Multi-Genre Broadcast ASRU challenge task (MGB1) \cite{Bell2015}. The full audio data consists of seven weeks of BBC TV programmes covering a wide range of genres, e.g. news, comedy, drama, sports, etc..  Systems were trained using a 200hr training set and a smaller 50hr subset. 
The data were collected by randomly sampling  audio  from 2,180 broadcast episodes. All utterances collected had a phone matched error rate of  $ < $ 20\% between the sub-titles and the  distributed MGB1 lightly supervised output.  The  official dev.sub set was employed as a validation set and consists of 5.5  hours of audio data  from across 12 episodes.  A separate evaluation set was used to  estimate the generalisation performance of each candidate model.  For our experiments  we took the remaining 35 shows from the MGB1 dev.full  and denote this set  as dev.sub2 for evaluation purposes.  The segmentations used for all experiments were taken from the reference transcriptions. Further details of the data preparation  are in \cite{Woodland2015}.

All experiments were conducted using the HTK 3.5 toolkit \cite{Zhang2015,HTKBook15} with supports DNN and HMM training including discriminative sequence training. This  work focuses on training hybrid DNN HMMs using traditional fully-connected feed-forward layers.  The input to the DNN was produced by splicing together  40  dimensional log-Mel filter bank (FBK) features extended with their delta coefficients across 9 frames to give a 720 dimensional input per frame.
These features were normalised at the utterance level for mean and at the show-segment level for variance  \cite{Woodland2015}. The DNN used an architecture of 5 hidden layers each with 1000 nodes with sigmoid activation functions.
The output softmax layer had context dependent phone targets formed by conventional decision tree context dependent state tying. 
The output layer  contained 4k/6k nodes for the 50h/200h training sets.

To make  fair comparisons between different optimisation frameworks all models were trained using lattice-based MPE training. Prior to sequence training, the DNN model parameters were initialised using frame-level CE training. 
To check the efficacy of each optimisation method, decoding  was performed  on the validation set at intermediate stages of training using the same weak pruned biased LM that we initially used to create the initial MPE lattices. Having such  a framework is advantageous as it also allows us to investigate over-fitting due to training criterion mismatch between the MPE criterion and the WER.

\subsection{Training configuration for SGD}
From  initial experimental runs, it was found that the best learning rate was $1 \times10^{-4}$.  As an optimiser, SGD on its own can be fairly unstable with network topologies where the gradient has to propagate through many layers. To stabilise training, sequence training with SGD was  accompanied by layer dependent gradient clipping  to prevent  network saturation and  ensure smooth propagation of gradients during the training process.

\subsection{Training configuration for NG \& HF-variants}
From preliminary experiments it was found that using batch sizes of roughly 25 hrs  gave a good balance between the number of updates and using good gradient estimate. When  the number of CG iterations is limited, initialising CG with a good estimate  of the gradient is desirable, as within a few iterations a good descent direction can be found. However this comes at the cost of fewer HF/NG updates per epoch as  larger batch sizes are needed to reduce the variance associated with the gradient estimates.  In  preliminary tests, it was found that  increasing the batch sizes beyond 25hrs yielded no significant improvements. To make up the NG/HF minibatch, we followed  Kingsbury's approach \cite{Kingsbury2012} and sampled only  1\% of the training set.

\section{Experiments} \label{sec:results}
\subsection{Experiments on 50hr MGB1 dataset}
On the 50hr setup, we investigated the efficacy of  the proposed optimisation framework and compared to both SGD and HF. Table~\ref{tab:1} summarises the results.

\begin{table}[H]
\centering
\tabcolsep=0.12cm
\begin{tabular}{|c|c|c|cc|c|c|}
\hline
method & \#epochs & \#updates &\multicolumn{2}{c|}{phone acc.} & WER\\
& & &  train & dev.sub & dev.sub \\
\hline
CE  & N/A & N/A & 0.7988 & 0.6549 &45.2 \\
 \hline\hline
SGD & 8 &2.62 $ \times 10^5$& 0.8476 & 0.7044 & 42.0\\
\hline
  HF & 40 & 80 &  0.8461 & 0.6967 & 42.2\\
 \hline
 NG & 30 & 60& 0.8677   &  0.7089 &41.6\\
 \hline
\end{tabular}
\caption{\label{tab:1} Sequence training and dev.sub MPE phone accuracy/WER for the 50hr training set. The WERs shown at the last column were computed using the weak pruned biased LM used in MPE training.}
\end{table}

It can be seen from Table~\ref{tab:1} that of the sequence training approaches the NG method produces the highest training and validation (dev.sub) MPE phone accuracies and the lowest WER using the MPE small biased LM. The NG approach
replaces the  `Hessian' component of the HF optimiser with the empirical FI (FI) matrix and this leads to both  faster and improved convergence which makes the proposed optimisation framework  more effective 
where only a relatively few updates are performed with large batch sizes. The NG optimiser requires just 60 updates to both better model the training data and improve the WER on the validation set\footnote{These gains were found to be consistent when we re-ran training with different sampling schemes.}.  In comparison to SGD baseline, both HF and NG training were observed to be highly stable  and did not require any form of additional update clipping.  This is attributed to the fact that at each iteration, the direction explored is a modified gradient estimate that has been appropriately rescaled  using the local KL curvature information. 

While both the HF and NG optimisers use more training epochs than SGD, each epoch has fewer updates and is somewhat more efficient in constructing a geodesic in the parameter space. Furthermore since the batch gradient computations are  inherently data parallel,  a batch style optimisation framework such as HF/NG becomes time efficient in a distributed setting where the gradient calculations can be easily parallelised and still yield identical updates. The relative contribution to the total computational cost by the CG iterations were 15\% for HF and 18\% NG\footnote{The extra CG cost associated with NG is attributed to  doing more CG iterations. With HF, in our preliminary runs, running CG beyond 5 iterations was not found to yield any significant improvement in the quality of the updates. Since NG is effectively a first order method, we found that in its case we needed to do slightly more iterations at each run.}.  


To investigate whether the WER reductions still hold with a with  stronger LMs, additional decoding passes on the dev.sub validation set were run with 158k vocabulary  bigram and trigram LMs used in \cite{Woodland2015}:
\begin{table}[H]
\begin{adjustbox}{max width=0.5\textwidth}
\begin{tabular}{|c|c|c|c|c|}
\hline
LM  & ~SGD~ &  ~~HF~~ & ~~NG~~ \\
\hline
158k Bigram 	&   39.2 & 39.4 & 39.0 \\
\hline
158k Trigram 	& 32.8 & 33.0 & 32.6 \\
\hline
\end{tabular}
\end{adjustbox}
\centering
\caption{\label{tab:3} \%WER differences between different optimisers on dev.sub with 158k vocabulary  bigram/trigram LMs on dev.sub (50hr)}
\end{table}
From Table~\ref{tab:3}, it can be seen that the same trends on dev.sub continue with NG giving greater reductions in WER than models trained with either SGD or HF.

\subsection{Experiments on 200 hr MGB1 dataset}
The experiments on the 200 hr training set were performed to ensure that the sequence training techniques generalise to a somewhat larger training set (and output layer size) and also to present more detailed comparisons of how training proceeds. 
The use of the proposed NG  method was also compared to   the Dynamic Stochastic Average Gradient HF variant (DSAG-HF), presented in \cite{Dognin2013}.  Table~\ref{tab:4} provides a summary of the results while Fig.~\ref{fig1} compares the performance  at intermediate stages of training.

\begin{table}[H]
\tabcolsep=0.15cm
\centering
\begin{tabular}{|c|c|c|cc|c|c|}
\hline
method & \#epochs & \#updates &\multicolumn{2}{c|}{phone acc.} & WER\\
& & & train  & dev.sub & dev.sub \\
\hline
CE  & N/A & N/A &0.8106& 0.6986 & 41.2\\
 \hline\hline
SGD & 8 & 9.27 $ \times 10^5$& 0.8684 & 0.7601 & 38.2\\
\hline
  HF & 15 & 120 &  0.8417 &0.7365& 38.8 \\
 \hline
 DSAG-HF & 15 & 120 &0.8499 &0.7456& 38.5\\
 \hline
 NG &  15& 120 & 0.8601   & 0.7534& 37.9\\
 \hline

\end{tabular}

\caption{\label{tab:4}Performance achieved with different optimisers on the 200 hr setup.}

\end{table}

Table~\ref{tab:4} shows that as for the 50 hr case, sequence training produces increases in MPE phone accuracy and reductions in validation set WER for all optimisers. In this case a total of 8 epochs were performed with SGD and 15 epochs (120 updates)
with HF, DSAG-HF and NG. In DSAG-HF,  the HF optimiser is equipped with a `momentum' component which essentially contributes to initialising CG with  a weighted sum of  the current and previous gradient directions. In Fig.~\ref{fig1},  it can be observed that although such a modification improves optimisation of the MPE criterion, improvements achieved in the MPE criterion do not necessarily correlate with the model's ability to achieve lower WERs.  In contrast,  replacing the GN with the empirical FI matrix  leads to  progressively better updates at each epoch.  In Fig.~\ref{fig1},  the reductions in validation set WER are very similar over the first 8 epochs between NG  and SGD, but NG only requires eight updates per epoch. Furthermore, through  following a path where the MPE training criterion  remains a good approximation  to the WER, the NG method allows training to converge  to a  better solution than either SGD or any of the HF variants.

\begin{table}[H]
 \tabcolsep=0.08cm
 \begin{tabular}{|c|c|c|c|c|}
\hline
~~~LM~~~  & ~~SGD~~ &  ~~~HF~~~ & DSAG-HF& ~~~NG~~~ \\  
\hline
158k Bigram &35.0 & 35.3 &  35.2 & 34.7\\
\hline
158k Trigram & 29.3 & 29.3 &29.2   &29.0 \\
\hline
\end{tabular}
\centering
\caption{ \label{tab:5} \%WER differences between different optimisers on dev.sub with 158k vocabulary  bigram/trigram LMs on dev.sub (200 hr).}
\end{table}

The resulting DNN acoustic models were tested with stronger  LMs as for the  50hr setup setup and the results are shown in Table~\ref{tab:5}. It can be seen  again that the NG method results in lower WERs than SGD or the HF variants.

\begin{table}[H]
 \tabcolsep=0.08cm
 \begin{tabular}{|c|c|c|c|c|c|}
\hline
 LM & ~~CE~~ & ~~SGD~~ &~~~HF~~~ &DSAG-HF &~~~NG~~~ \\
 \hline
158k  Trigram &33.1 & 30.8 & 31.0 & 30.8 &30.5\\ 
 \hline
\end{tabular}
\centering
\caption{\label{tab:6} \%WER differences between different optimisers on  dev.sub2  with 158k trigram (200 hr)}
\end{table}

Finally, these DNNs were tested on the dev.sub2 set as an evaluation set  that was not used for setting training hyper-parameters to ensure that the trends observed above generalise. Table~\ref{tab:6} shows the WER results  using the 158k trigram model and shows that again that the model trained with NG achieves the largest reductions in the WER due to sequence training. Furthermore these improvements have been fairly consistent between the validation dev.sub and evaluation dev.sub2 test sets.
On average the NG method can be seen to provide approximately a 1\% relative reduction in WER over both the SGD baseline and the DSAG-HF variant. While this improvement it fairly small, it is consistent and  a statistical significance test (sign test of the  word error rates at the episode level) showed that the improvement due to NG over each of the other methods is highly statistically significant ($p < 0.001$).

\begin{figure}[t]
\hspace*{-0.6cm}\includegraphics[scale=0.45]{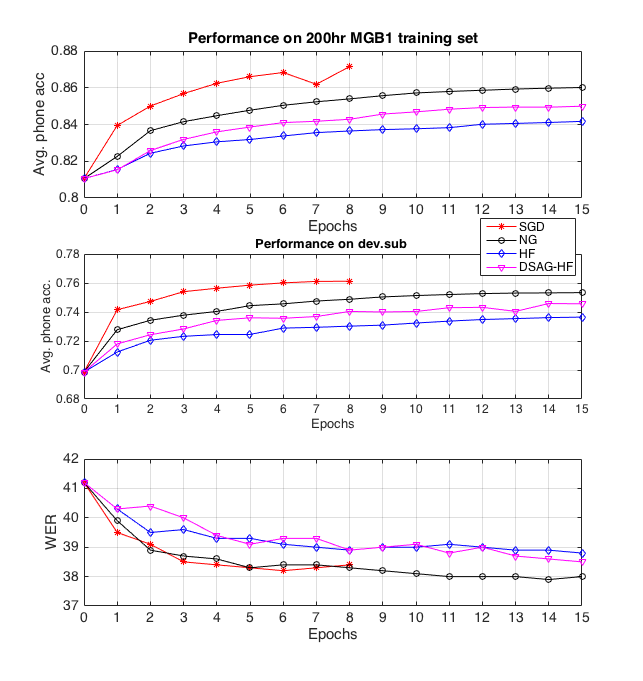}
\vspace*{-2em}
\caption{\label{fig1} The evolution of  the MPE phone accuracy criterion on the training and validation (dev.sub) sets (top 2 graphs). Also (lower graph) WER with MPE LM on dev.sub as training proceeds (200 hr).}
\end{figure}

\section{Conclusions}

This paper has described a discriminative sequence training method based on the NG approach and has shown that it can improve the optimisation performance of the discriminative sequence training objective functions used 
in speech recognition systems. This novel technique provides the same advantages as large batch  HF methods in terms of stability and and being inherently data parallel. However it converges more quickly and finds better performing final models.
It was evaluated using training and test data from the MGB1 transcription task where it was shown that as well as being effective in optimising the MPE objective function, it tends to generalise better and leads to reduced WERs on independent test data.
 Future work  will involve extending the proposed framework to sequence training of DNN architectures with recurrent topologies.

\label{sec:ref}


\end{document}